\documentclass[journal,comsoc]{IEEEtran}
\usepackage[T1]{fontenc}

\ifCLASSINFOpdf
\usepackage[pdftex]{graphicx}
\else
\usepackage[dvips]{graphicx}
\fi
\usepackage{amsmath}
\usepackage[cmintegrals]{newtxmath}
\usepackage{url}

\usepackage{placeins}

\usepackage{todonotes}

\begin{document}

© 2021 IEEE. Personal use of this material is permitted. Permission from IEEE must be obtained for all other uses, in any current or future media, including reprinting/republishing this material for advertising or promotional purposes, creating new collective works, for resale or redistribution to servers or lists, or reuse of any copyrighted component of this work in other works.
%
\title{Morpho-evolution with learning using a controller archive as an inheritance mechanism}
%
%
%

\author{Léni K. Le Goff, Edgar Buchanan, Emma Hart, Agoston E. Eiben, Wei Li, Matteo De Carlo, Alan F. Winfield, Matthew F. Hale, Robert Woolley, Mike Angus, Jon Timmis, Andy M. Tyrrell 
\thanks{L. K. Le Goff and E. Hart are with the School of Computing, Edinburgh Napier University, Scotland, UK, contact l.legoff2@napier.ac.uk}
\thanks{E. Buchanan, W. Li, R. Woolley, M. Angus, A. M. Tyrrell are with the Department
of Electronic Engineering, University of York, England, UK}
\thanks{M. F. Hale and A. F. Winfield are with Bristol Robotics Laboratory, University of the West of England, UK}
\thanks{J. Timmis is with School of Computer Science, University of Sunderland, England, UK}
\thanks{M. De Carlo and A. E. Eiben are with Department of Computer Science, Vrije Universiteit Amsterdam, NL}}

%
%

\markboth{IEEE TCDS 2021 Special Issue on Towards autonomous evolution, (re)production and learning in robotic eco-systems}
{Shell \MakeLowercase{\textit{et al.}}: Bare Demo of IEEEtran.cls for IEEE Communications Society Journals}
%



\maketitle


\begin{abstract}

The joint optimisation of body-plan and control via evolutionary processes can be challenging in rich morphological spaces in which offspring can have body-plans that are very different from either of their parents. This causes a potential mismatch between the structure of an inherited controller and the new body.
To address this, we propose a framework that combines an evolutionary algorithm to generate body-plans and a learning algorithm to optimise the parameters of a neural controller. The topology of this controller is created once the body-plan of each offspring body-plan is generated. The key novelty of the approach is to add an external archive for storing learned controllers that map to explicit  `types' of robots (where this is defined with respect the features of the body-plan). By learning from a controller with an appropriate structure inherited from the archive, rather
than from a randomly initialised one,  we show that both the speed and magnitude of learning increases over time when compared to an approach that starts from scratch, using two tasks and three environments.  The framework also provides  new  insights  into  the  complex interactions  between  evolution  and  learning.
\end{abstract}


%
\IEEEpeerreviewmaketitle

\section{Introduction}

    

The idea of embodied intelligence --- describing the design and behaviours of physical objects situated in the real-world --- was first introduced by Brooks in 1991 \cite{brooks91}. Pfeifer and Bongard's seminal text "How the body shapes the way we think” \cite{pfeifer2006body} expanded on the idea that intelligent control is not only dependent on the brain, but at the same time both constrained and enabled by the body.  Increasingly, artificial evolution approaches have been used in robotics  to jointly optimise both the body-plan and controller of a robot to accomplish a desired task. This has the potential advantage of allowing evolution to discover the appropriate balance between morphological and brain complexity and functionality.
 
However, much of this work has taken place in restricted morphological spaces, for example using regular shaped  modules to construct body-plans, in which each module can be individually actuated \cite{Kriegman1853, iida}. In richer spaces which can give rise to complex and irregular robot skeletons with multiple forms of sensing and actuation (e.g. joints and/or wheels), then more complex controllers  that link  multiple sensors and actuators  are required. In addition, the evolutionary process becomes more challenging: reproduction between two morphologically distinct parents might result in a viable body-plan, but a directly inherited controller is  at best unlikely to provide adequate control and, at worst, will not work at all because inputs and outputs do not correspond to the new body-plan.

 One approach to address this is to evolve a morphology-independent control mechanism, for example using a compositional pattern producing network (CPPN) \cite{stanley2007compositional} to \textit{generate} a  controller, thereby enabling direct inheritance of the generator \cite{le2020pros}. However, EAs using generative methods tend to need more generations to converge than EAs using fixed size genomes. 
An alternative is to add a learning cycle into the evolutionary loop \cite{eiben2013triangle,EibenHart}. This can either improve an inherited controller over an individual's lifetime -- when the inherited controller has an appropriate structure -- or learn a new controller from scratch otherwise.  Here, we follow the latter approach and propose a novel framework  for combining evolution and learning that is capable of joint optimisation of body and control of robots in a complex morphological space when using controller encodings that do not permit directly inheritence,  i.e. when the topology of a child controller does not match the inherited body. 
 
The framework contains a morpho-evolutionary algorithm (MEA)  to optimise the body-plan and a learning algorithm to optimise the parameters of the controller. Two optimisation processes are nested: for each body-plan produced  with the MEA,  the learning process is invoked to optimise its controller.  The key novelty of the approach is the addition of an external {\em controller archive}: this multi-dimensional archive stores  the best found controller for a given 'type' of robot, where \textit{type} is defined by a vector describing the robot's morphological features (e.g. number of wheels, number of sensors of Type A, number of sensors of Type B, etc.).  If a body-plan is produced that is of the same type as a controller already stored in the archive, the learning process is initiated with this controller, otherwise it starts from scratch. The archive is updated over the generations as better controllers are found. Essentially the archive can be viewed as a form of inheritance, storing successful controllers per robot type that can be used to bootstrap \textit{learning }in future generations. 
Hence the framework is named MELAI: \textit{morpho-evolution with learning using archive inheritance}. Specifically, for the MEA,  we use the matrix-based CPPN morpho-evolution (MCME) introduced in our previous work \cite{buchanan2020bootstrapping} to evolve body-plans. The learning algorithm used is a novelty-driven evolution-strategy, that uses an increasing population size (NIP-ES), and was also introduced in our previous work \cite{le2020sample}. It learns the weights of an controller specified by an Elman network that has a topology matching the generated child body-plan.


The contributions of the method are two-fold: (1) it offers a novel approach for the joint optimisation of both body-plans and controllers of robots, that integrates evolution and learning: uniquely, it uses a morpho-evolutionary algorithm for the former and an evolution-strategy for the latter; (2) it proposes the use of an external archive as an efficient mechanism for transferring control knowledge from parents to offspring in situations where offspring are morphologically distinct from their parents. 

We show the benefits of using an archive (which interacts only with the learning process) 
in terms of increasing the efficiency of the approach (compared to methods that learn from scratch) and provide new insights into the interplay of evolutionary and learning processes. Moreover, as an additional contribution, our results show the emergence of different types of robots for different tasks.




The rest of the paper is organized as follows: section~\ref{sec:rw} analyses previous studies related to  the joint optimisation of robot body-plans/controllers, then in section~\ref{sec:methods} MELAI is explained in detail; the experimental protocol is described in section~\ref{sec:exp} and the results are presented in section~\ref{sec:results}.
Finally, sections~\ref{sec:disc} and \ref{sec:concl} discuss the results and conclude the paper.

\section{Related Work}\label{sec:rw}

Among the numerous studies in the field of evolutionary robotics, the majority address either the evolution of the body-plan alone \textit{or} the evolution of the controller alone. This section focuses on literature which describes methods for the \textit{joint} optimisation of body-plans and controllers. We focus attention on approaches that permit the evolution of offspring that require a controller topology that is different to either parent. That is, methods which evolve changes to morphology that do not impact the topology of controller (i.e. the number of inputs and outputs of a neural controller) are out of scope. For example, this excludes work in which morphological change is restricted to repositioning sensors \cite{buason2005brains} or altering the length, weight and size of leg-joints, e.g.  \cite{nygaard2017overcoming,nygaard2018real,nygaard2020environmental, juarez1998design}.
We first discuss methods that create  controllers that are directly correlated to a specific type of body-plan, followed by morphology-independent methods, i.e. those that are capable of \textit{generating} a controller for any given body-plan.

A  naïve approach to avoiding a potential mismatch between a controller and body-plan is to evolve only the body-plan and then learn a new controller with the correct topology from scratch for each child body-plan. The work of Gupta \textit{et al} \cite{gupta2021embodied} follows this approach by using evolution for the body-plans and reinforcement learning to optimize the controllers for simulated robots composed from articulated 3D rigid parts connected via motor actuated hinge joints. Learning starts from a randomly initialised controller for each body-plan, and uses a distributed implementation across  multiple CPU to minimise computational cost.
Lia et al. \cite{liao2019data} also proposed a similar nested optimisation process, with the aim of finding the best morphology for a walker micro-robot. 
Here, Bayesian optimisation is used to learn the  controller. 
However, despite Bayesian optimisation being known to be sample efficient, it only works well for small parameter spaces \cite{le2020sample}.

Instead of learning from scratch, an alternative approach is to use a morphology-independent control generation mechanism that can generate control parameters for any given body-plan.
For example, Cheney \textit{et al} \cite{cheney2016difficulty,cheney2018scalable} evolved soft-robots built from voxels, in which each voxel has a parameterised local controller. Both body-plans and controller parameters are outputs of two separate compositional pattern producing networks (CPPN) \cite{stanley2007compositional}, both of which was evolved via the well known neuro-evolution with augmenting topology (NEAT) algorithm\cite{stanley2002evolving}. However, due the distributed nature of the controller, the variety of possible behaviours for the same body-plan is limited.
Sims \cite{sims1994evolving} also used a decentralised form of control. They proposed a genotype that contains a nested graph: the graph specifies morphological nodes describing the robot shape, each of which contains another graph specifying the neural circuitry for that node. More recent work has achieved a similar effect with the use of Lindenmayer-Systems (L-systems) decoding instead of a graph-representation, e.g.  \cite{hornby2003generative, miras2018search, miras2020evolving}.

In the latter work \cite{miras2020evolving}, an additional learning mechanism was applied to improve the inherited brains of newborn robots: the authors showed that learning not only influences the morphology of the resulting robots, but that also, the capacity to learn increases over generations. Jelisavcic \textit{et al.} \cite{jelisavcic2019lamarckian} also employed a learning mechanism. Their genome carried a pool of CPPNs which can be used to specify the weights of a controller generated to match the child body-plan. Differently to the work of Cheney \textit{et al}  which encoded a single CPPN that undergoes evolution, here a child inherited a subset of CPPNs from each parent, then a learning algorithm (HyperNEAT) was applied to the inherited pool to evolve a new pool. The process was thus Lamarckian.

To summarise, in the context of combined optimisation of body and control, on the one hand the literature has shown that using generative encodings (with and without additional learning) can mitigate the issues arising regarding inheritance of controllers that might not be applicable to a new child body-plan. However, these methods often require many evaluations to converge \cite{le2020pros} and add additional hyper-parameters which may be difficult to optimise. On the other hand, neural controller encodings which are explicitly tied to a body-plan can be rapidly optimised as they only require weight optimisation rather than topology. Although they often cannot be inherited, this can be addressed by learning a controller from scratch, e.g. as in \cite{gupta2021embodied}, although at the expense of ignoring any previously learned knowledge.

In this paper, we choose to use a fixed structure neural network for reasons of efficiency, motivated by the goal of eventually evolving directly in hardware. As in previous works, we use a learning algorithm to optimise a controller that has a fixed structure that matches the new body-plan \cite{le2020sample}.  However, in order to avoid starting from scratch for each body-plan as in previous work, we introduce a novel method for storing past solutions that can be accessed by the learning algorithm to bootstrap learning. This takes the form of an archive that stores the weights of a controller for each \textit{`type'} of robot that has previously been encountered as described in the previous section.  This archive or `brain pool' is dynamically composed and adapted during the evolutionary process.

Note that the term \textit{archive} should not be confused with other uses of the word in the wider evolutionary literature. For example, archives are commonly used in multi-objective optimisation to either drive the population toward the Pareto front or to maintain the population diversity, that is, they directly interact with an evolving population. Our approach has more in common with methods which try and enhance a search process by re-using past experience gained when solving related problems. For example, Louis and McDonnell \cite{louis2004learning} maintained a store of past solutions from similar instances which are periodically injected into an evolving population. However their approach is only applicable if instances share structural properties, hence cannot be  applied to controllers with different topologies. Feng \textit{et al} \cite{feng2015memes} attempted to reuse structured common knowledge captured in the optimized solutions of past search experiences in a form independent of solution representation, however their specific implementation was tailored to combinatorial optimisation. Here, we draw inspiration from \cite{louis2004learning} in maintaining an archive of past solutions. Extending this concept, we store solutions with different \textit{types}, corresponding to different controller topologies, organised in the form of a grid. While our use of a grid-based container to organise solutions according to set of features resembles the quality-diversity algorithm MAP-Elites \cite{mouret2015illuminating}, the role of the archive in our work is different: (1) in MAP-Elites, the grid is first filled using a randomly generated solutions, which form a \textit{population}; (2) the population is acted on by an evolutionary process that performs mutation and/or crossover to generate new solutions which are projected back to the grid. In our work, the archive does not act as a population but as a library. Cells are only filled with a controller if and when a particular `type' is generated by the MEA. A single controller from the library is selected to initialise a separate learning algorithm. In summary, the processing of filling cells is governed by the MEA, while the stored controllers are used to inform the learning algorithm.


\begin{figure}[h]
\begin{center}
\includegraphics[width=\linewidth]{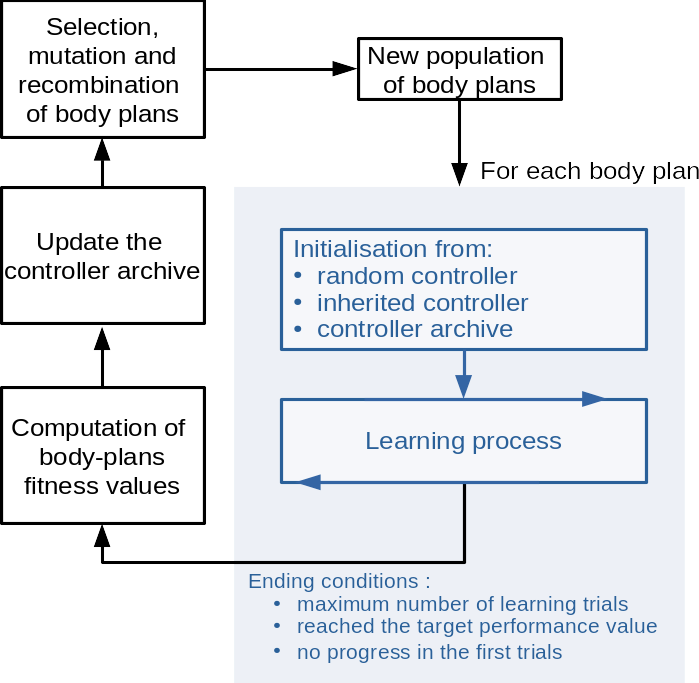} 
\end{center}
\caption{Diagram illustrating the MELAI algorithm. MELAI has two nested optimisation processes. As main process, a morpho-evolution algorithm, shown in black, divided in four main steps: \emph{computation} of the fitness values, \emph{update} the controller archive with the best ones from the current population, \emph{Selection, mutation, and recombination}, and finally send the new population of new body-plans to the learning process. For each body-plans, a controller is learned, shown in blue. The learning process can either start from a random controller, a inherited controller, or a controller from the archive. Then, the learning process run until reaching an ending condition.} 
\label{fig:mnipes}
\end{figure}



\section{Methods}\label{sec:methods}

\subsection{Algorithm Description}

Morpho-evolution with learning using an archive inheritance (MELAI) is a nested optimisation algorithm. 
As illustrated in figure~\ref{fig:mnipes}, body-plans are optimised with an evolutionary algorithm, then for each body-plan, a learning process is used to optimise their controller. 

The first optimisation algorithm (a morpho-evolution algorithm MEA) uses a generative encoding to produce the robot's body-plan, based on our previous work described in \cite{buchanan2020bootstrapping}. This is a matrix-based CPPN morpho-evolution denoted MCME. 
The second optimization algorithm (learning) optimises the parameters of a controller which is a fixed size neural network structure. Therefore, the number of parameters is fixed. The novelty-driven increasing population evolutionary strategies (NIP-ES) algorithm \cite{le2020sample} is used for learning. 
A detailed description of both MCME and NIP-ES  is given in the Supplementary Materials. Although the instantiation of MELAI described in this paper uses MCME and NIP-ES, the framework itself is general in that any kind of MEA or learning algorithm could be used.

Inheritance of controllers from parents to children is challenging for MELAI as previously noted, since children might have different body-plan configurations than their parents. One way to address this issue is to learn the controller for each robot from scratch as in \cite{gupta2021embodied}. However, this has a number of disadvantages, including the fact that previously learned information from past learning cycles is wasted. In order to address this issue MELAI facilitates three initialisation options: 

\begin{enumerate}
    \item Select a controller from the archive with the same number of sensors and actuators assuming one exists.
    \item Start from a randomly initialised controller.
    \item Direct controller inheritance if  the parent and child share the same number and type of actuators and sensors. 
\end{enumerate}

In this paper, the third option of direct inheritance is not considered because the encoding and the morphological space used in MCME  make it unlikely that a parent and child will share the same number and type of actuators and sensors. Thus, the benefit of direct inheritance is likely to be negligible. This enables the experiments to focus directly on determining the benefit of the archive over random initialisation.

As noted, the learning algorithm used is NIP-ES,  first described in  \cite{le2020sample}. The core of this method is  a co-variance matrix adaptation evolutionary strategy (CMA-ES) algorithm in which a normal multivariate distribution (MVND) is used to sample a new population at each iteration. When using a controller from the archive, it is used to provide  the starting mean of the MVND and thus the starting population is sampled in the surrounding of the parameters of this controller. When starting from scratch, CMA-ES starts from a random mean. 

The learning process stops when one of the ending conditions is reached:
\begin{itemize}
    \item \emph{A satisfactory solution is reached}: the learning algorithm finds a controller with a task-performance value  above a certain threshold.
    \item \emph{The maximum number of evaluations is reached}: each optimisation process has a maximum number of updates. For the MEA, this parameter is the number of generations and for the learning process is the number of evaluations. The values of these parameters have to be chosen according to the difficulty of the task and environment, but also according to the constraints of the system on which the algorithm is running. In this study, a constraint of 100000 maximum evaluations is used.  Given this overall budget, an additional choice that must be made is to decide how to divide it between the MEA and the learning process.
    \item \emph{The performance of the robot stays very low during a trial period.} The trial period is defined by a fixed number of evaluations (50 in all experiments). If a robot has not moved (i.e. very low performance) by the end of this period then the  learning process stops.
\end{itemize}

In the rest of the paper, the \emph{fitness} indicates the value used by the MEA for selection.
The term \emph{task-performance} is used by the learning algorithm to assess the quality of a behaviour.  Several fitness functions can be used for the MEA. The most natural fitness function to use will be the \textit{best} task-performance value found during the learning. This is the one used in this paper. However, the learning process produces additional data which could also be exploited in future, such as statistical information regarding the progress of the learning algorithm or the novelty-scores used by NIP-ES.

\begin{figure}[h]
\begin{center}
\includegraphics[width=\linewidth]{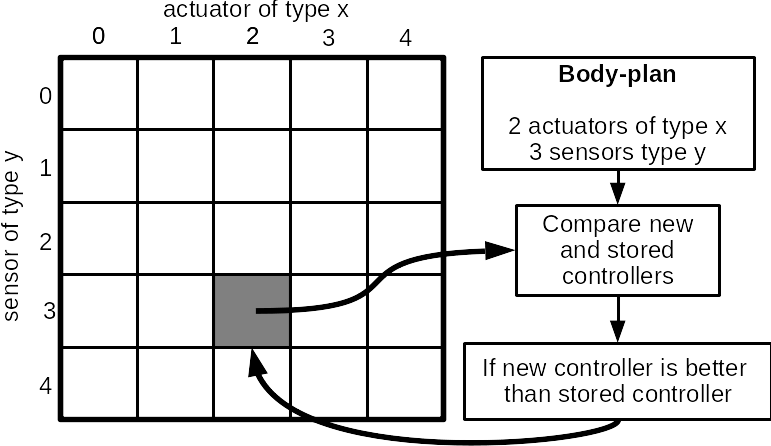} 
\end{center}
\caption{Diagram illustrating the update of the controller archive. A body-plan with 2 actuators of type x and 3 sensors of type y has a new controller output of the learning process. If the cell corresponding to 2 actuators of type x and 3 sensors of type y is not empty, the new controller is compared with the stored one. The new controller replace the stored one  if its task-performance is greater.} 
\label{fig:ctrlarch}
\end{figure}

\subsection{Controller Archive}

The controller archive stores the best controller found for different `types' of robot. A `type' is defined as a $n$-tuple, 
 where each dimension represents a component of the body-plan, e.g. a specific form of sensor or actuator. An archive thus consists of an n-dimensional cube, where each dimension is discretised into cells corresponding to the number of each kind of component. A cell contains the best controller found for a body-plan described by a given tuple generated by the MEA, and is empty otherwise.

For instance, let us consider a morphological space with  one kind of actuator (x) and sensor (y) as shown in figure~\ref{fig:ctrlarch}. Assume a body-plan is generated with  2 actuators and 3 sensors. When the learning process has ended, its controller's task-performance value is compared with the stored one in the corresponding cell and replaces it if its task-performance value is greater. If the cell is empty the new controller is added to the cell.



We consider two kinds of actuator (wheel and joint), and one type of sensor. So, a 'type' of robot in this case is defined by a tuple <num\_sensors, num\_wheels, num\_joints>.  The controller archive can be considered as a new form of inheritance. All the behavioural knowledge from  past generations is stored in a common archive to be used by future generations. In this way, new child robots can leverage the learned behaviours of their ancestors.

\section{Experiments}\label{sec:exp}

\subsection{Experimental protocol}

The experiments presented in this article aim to answer the following questions: 
\begin{itemize}
    \item[\textbf{1.}] To what extent does using a controller-archive to bootstrap learning 
    improve effectiveness and efficiency when compared to learning from scratch?
  \item[\textbf{2.}] To what extent does the distribution of effort between the morpho-evolution process and the learning process influence performance?


    
\end{itemize}


Experiments are conducted with and without the controller archive to answer question \textbf{1.} In this way, the role of the controller archive in MELAI can be isolated. In the results section, the variant of MELAI that does not use a controller archive is denoted morpho-evolution with learning (MEL). 

Two tasks are used for these experiments: exploration and photo-taxis with multiple targets. They are explained in detailed in section~\ref{sec:env}. For both task, the learning process has a budget of 200 evaluations. This learning budget is optimal for our experiments (see figure~\ref{fig:evalvar}). For the exploration task, the evolution runs for 20 generations and for the photo-taxis task it runs for 15 generations. Experiments on the photo-taxis task runs for less generation because each robot needs to be evaluated for each target, thus, it is computationally expensive.

To answer question \textbf{2.}, more experiments are conducted only on the exploration task. For all of them, a fixed budget of 100000 evaluations is shared between the two optimisation processes. This has the objective of studying any trade-offs in resource allocation between the two components of the framework. Parameter values tested are the following : [100,40], [150,30], [200,20], [400,10], [800,5], where the first value corresponds to the number of evaluations for each body-plan during the learning phase and the second to the number of generations of the MEA. Also, these variants of MELAI are compared with a baseline algorithm in which the learning process is replaced by a process that simply generates controllers using Latin Hypercube Sampling (LHS)\footnote{LHS samples evenly the parameters space. This gives a better sampling than using a simple uniform distribution.}\cite{mckay2000comparison}, i.e there is no directed learning, only random sampling of controllers. This variant is called Morpho-Evolution with Latin Hypercube Sampling (MELHS). MELHS runs for 40 generations and for each body-plans 100 random controllers are sampled.  These experiments are conducted only with the controller archive.


All experiments feature a population of 25 body-plans for the MEA.
The hyper-parameters used for the experiments are given in the supplementary materials.
All experiments are conducted in the three environments described in section~\ref{sec:env} and shown in figure~\ref{fig:env}.  Twenty replicates are performed for each experiment. The robots and environments are simulated using the CoppeliaSim version 4.2.0 simulator\footnote{https://www.coppeliarobotics.com/} The source code to run these experiments and their data are available here : \textit{the code and data will be provided if the paper is accepted.}

\begin{figure}[h]
\begin{center}\includegraphics[width=0.75\linewidth]{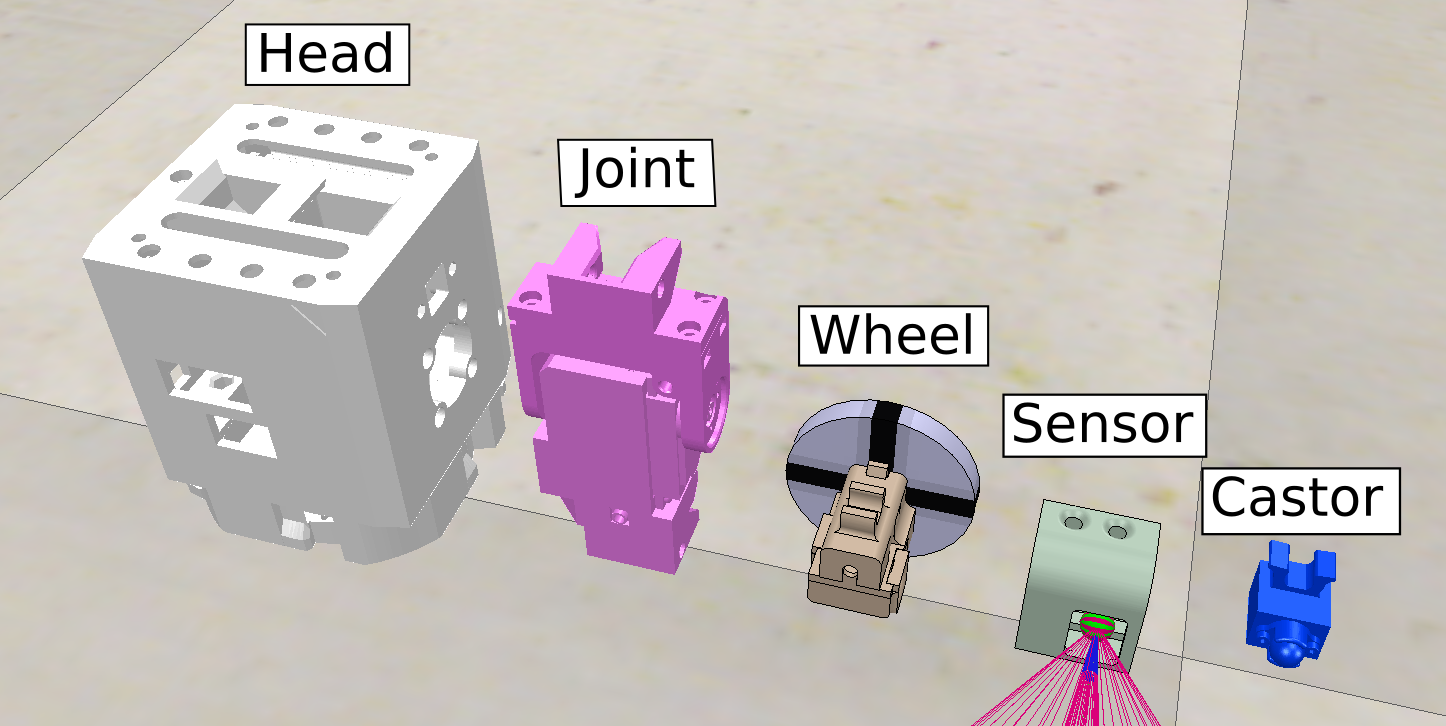} 
\end{center}
\caption{Active and passive components used for the experiments shown in this paper. The active components are the wheel, joint and sensor. The passive component is the caster wheel. The head in the central processing unit of the robot.} 
\label{fig:organs}
\end{figure}

\subsection{Body-plans}\label{sec:bodyplan}

The body-plans evolved using the MEA (as described in section\ref{sec:mcme})  have two main features:  skeletons can have complex and widely differing shapes and different numbers of components\footnote{All components have been designed to match the physical ones which  used in the ARE project \cite{hale2020hardware} in order to be able to conduct experiments directly in hardware in the future}. The components can be active (e.g. a wheel driven by motors, sensors) or passive (e.g. a caster-wheel) where the active ones interact with the controller. The different component types are shown in Figure~\ref{fig:organs}.
The controller  interacts differently with each active component:
\begin{itemize}
    \item Each \emph{wheel} takes one output from the controller which is translated to the speed of the rotational movement of the wheel.
    \item Each \emph{joint} takes one output from the controller which is  translated to the  frequency of the oscillatory movement of the joint. Oscillatory control has previously been demonstrated to give good results for locomotion \cite{jelisavcic2019lamarckian,miras2020evolving}. 
    \item Each \emph{sensor} provides two inputs to the controller where the first input is binary for the detection of a beacon and the second input is the distance from the closest obstacle. The detection of a beacon uses a simulated IR sensor and the distance measure uses a simulated time-of-flight sensor.
\end{itemize}

The head is the central processing unit of the robot and therefore is a special component of a body-plan. It is always positioned in the center of the skeleton (see figure~\ref{fig:robot_examples}). To obtain an accurate simulation, the mass of the body-plan has to be estimated. This is calculated as the sum of its components weights and of its skeleton. To estimate the mass of the skeleton, which  can take  various shapes, the density of a common plastic used for 3d printing is used to estimate the mass of a voxel. Skeleton mass is then obtained by multiplying the voxel mass by the number of voxels used. This results in robots that have mass between 0.5kg and 2kg depending on the size of the skeleton.

\subsection{Controllers}

The controller used in this study is a modified version of an Elman network \cite{elman1990finding}. An Elman network is a recurrent neural network with two hidden layers (see the figure in section C in Appendix II of the supplementary materials). The first hidden layer is fully connected to the input and output layers. Then, each neuron is forward connected to one neuron of the second hidden layer, called the context layer.  Neurons in the context layer (context units) are recursively connected to themselves, and the context units are also fully backward connected to the hidden layer. Each neuron has a sigmoid function as their activation function. The context layer acts like a short term memory and allows the network to process real number sequences such as time-series \cite{elman1990finding}. Elman networks have been shown to be more efficient as controllers for a navigation task than a simple feed forward network \cite{le2020pros} due to their ability to capture time-dependent information from sensors.

For each body-plan, each Elman network has a number of inputs and outputs corresponding to its body-plans number of sensors and actuators. The hidden and context layers have a fixed structure for all the body-plans. Transferring a trained network from one body-plan to another is not possible unless they have the same number of sensors and actuators.

\subsection{Task and environments}\label{sec:env}

Two tasks are used in the experiments presented in this paper: exploration and photo-taxis.

\begin{figure}[h]
\begin{center}\includegraphics[width=0.9\linewidth]{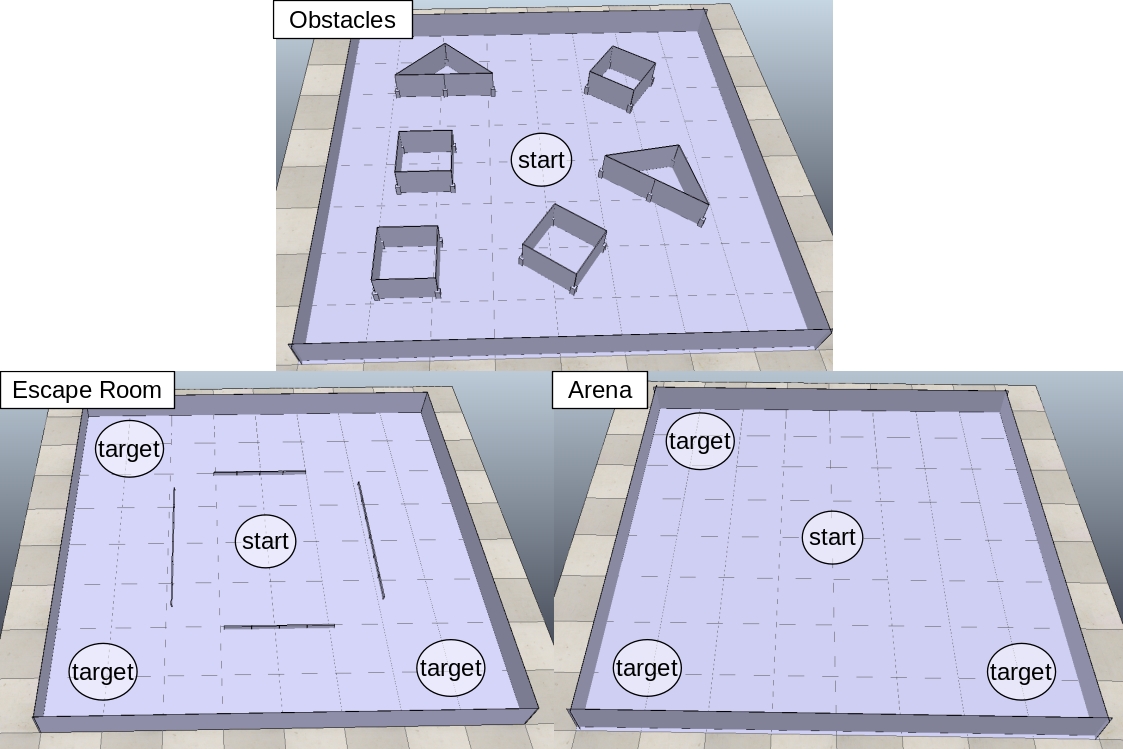} 
\end{center}
\caption{The three environments used for the experiments in this paper: obstacles, escape room and arena.} 
\label{fig:env}
\end{figure}

\paragraph{Exploration task}
In the \emph{exploration task}, the robot has to visit the most zones in a limited time. The zones are equal-sized squares forming a grid. The task-performance is computed by counting the number of zones visited and dividing the count by the total number of zones. The grid is 8 by 8 with cells of 25 cm sides, so the total number of zones is 64. The evaluation lasts 30 seconds and takes place in the \emph{obstacles} environment (see figure~\ref{fig:env}). In this task, there is not any target performance value therefore the first stopping criteria defined in section \ref{sec:methods} does not apply.

\paragraph{Photo-taxis task}
In the \emph{photo-taxis task}, the robot starts at one point and has to reach a target where a beacon is placed. The robot has first to find the beacon in the arena and then go toward it. As the beacon is detected using a simulated IR sensor, the robot can not see it when it is occluded by an obstacle. The robot is evaluated three times with the target at a different positions. The task performance is then the average of the task performance obtained in each evaluation.  
The task-performance function is the normalised distance between the final position ($p_f$) of the robot (at the end of the evaluation) and the position of the beacon ($p_b$) (see equation~\ref{eq:fitfct}). This distance is subtracted from 1 in order to define a maximisation function. The distance is normalized by the length of the diagonal of the arena. As the arenas are squares of two by two metres the diagonal measures $D = \sqrt{2^3} \simeq 2.83$

\begin{equation}\label{eq:fitfct}
    F = 1 - \frac{\|p_f - p_b\|}{D}
\end{equation}

The success threshold used to stop the learning process is equal to 0.95 for this task. This value correspond to a circle with a radius of 14 cm around the target.

\paragraph{Environments}
Three different environments (figure~\ref{fig:env}) are used in the experiments dubbed \textit{obstacles}, \textit{escape room} and \textit{arena}. They are all square (2 metres sides) and have a tiled floor. Each is  designed to be reproducible in reality. The tiles are spaced with a small gap of 1 millimetre which corresponds to the floor of our real arenas and the walls in the obstacles environments have feet to hold them standing. These constraints in design introduce small irregularities.

\begin{figure*}[t]
    \centering
    \includegraphics[width=0.32\textwidth]{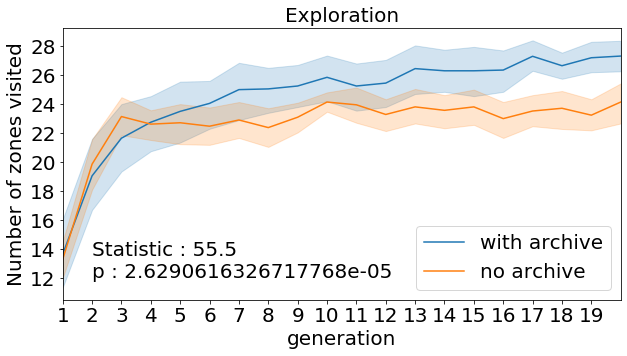}
    \includegraphics[width=0.32\textwidth]{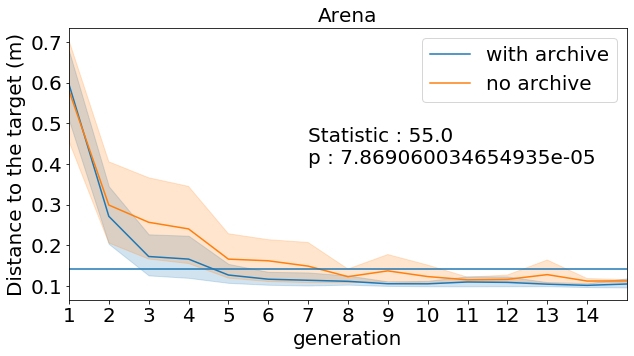} 
    \includegraphics[width=0.32\textwidth]{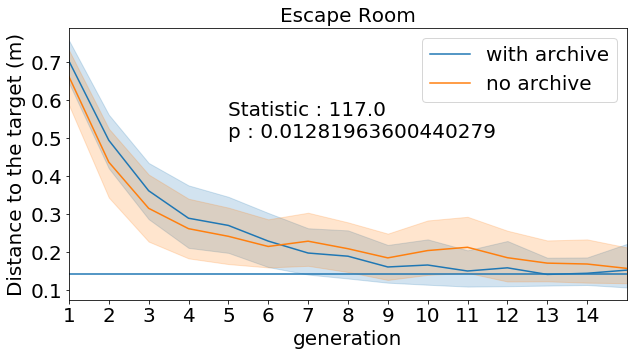} \\
    \includegraphics[width=0.32\textwidth]{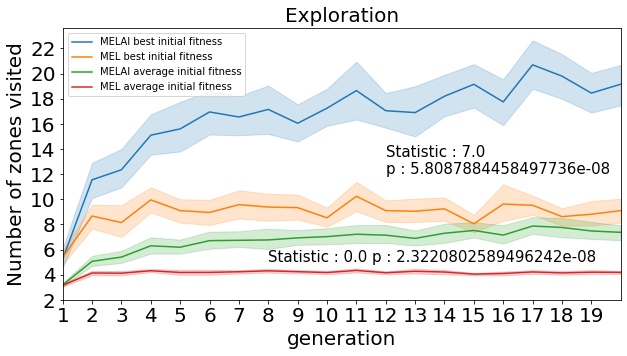}
    \includegraphics[width=0.32\textwidth]{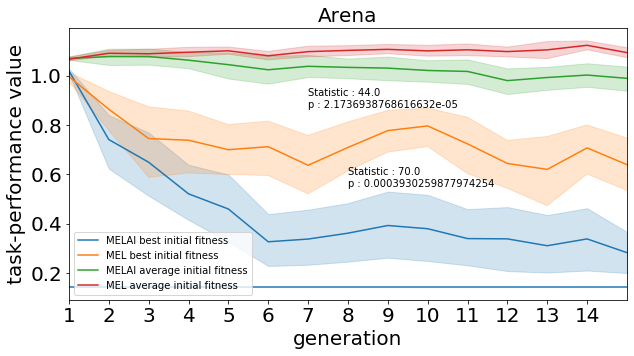} 
    \includegraphics[width=0.32\textwidth]{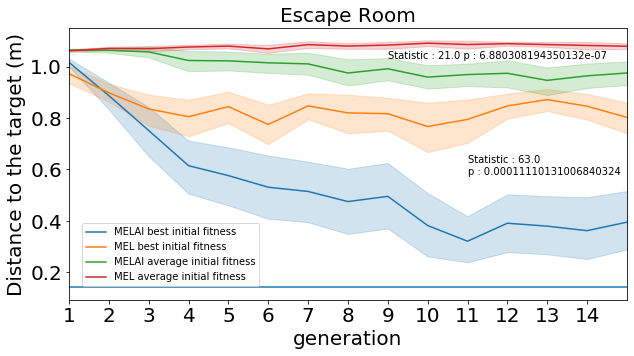} \\
    \caption{Measures of effectiveness of the algorithms. In the first row, the best fitness values over the generations and in the second row the best and average of the\textit{ initial} task-performance values over the generations. These experiments have been conducted with a budget of 200 evaluations per body-plans and 20 generations for the exploration task and 15 for the photo-taxis task The coloured areas correspond to the confidence interval around the mean. Difference between the distributions of the last generation is significant when a p-value and its critical value is indicated. The significant test is the Mann-Whitney U test.}
    \label{fig:fitness}
\end{figure*}
\section{Results}\label{sec:results}

This section is split in four parts: experiments related to the efficiency and effectiveness of the algorithm, the controller archive dynamics, the influence of learning, and the robots diversity. The first part focuses on comparing the algorithm with and without an archive by measuring the quality of the solutions produced and the efficiency of both variants. The second part examines the controllers stored in the archive in terms of their number and quality. The third part studies the influence of learning by comparing MELAI with different learning budget and a variant without learning. Finally, the last part investigates the influence of the task and environment over the type of robots generated. Where it is relevant a statistical test is conducted. The test used is the Mann-Whitney U  \cite{mann1947test} under the null hypothesis. For the series over the generations or the number of evaluations the distributions of the last generations is tested. All the experiments have been replicated 20 times.

\subsection{Efficiency and effectiveness}

Three measures are used to assess the efficiency and effectiveness of MELAI: (a) the best fitness for each population of MEA, (b) the best and average initial task-performance of the learning process of the population  (c) total number of evaluations. The best fitness (a) is calculated for each population after the learning has finished. The initial task-performance (b) is the lowest task-performance from the first iteration of NIP-ES.  The number of evaluations (c) used during one generation is the sum of the number of evaluations used by the learning process for each body-plan in the population. 
The best fitness and the initial task-performance is a measure of the performance (effectiveness) of the complete MELAI algorithm while the number of evaluations measures the efficiency of the algorithm.

Figure~\ref{fig:fitness} shows the plots of the best fitness (first row) and the best and average initial task-performance (second row) over the generations. For the exploration task and the arena with three targets, MELAI achieves better performances than MEL, which corresponds to more zones visited during the exploration task and smaller distance from the target for the photo-taxis task. With the exploration task, MELAI generates robots that are able to visit between 26 and 28 zones out of a potential 64, where with MEL, the best robots visits between 22 and 24 zones. In the arena, MELAI finds a solution which reaches the success threshold (approximately 0.14 meter from the target) in fewer generations ($~$ 4 generations) than MEL ($~6$ generations). All replicates find a solution that reaches the target threshold with MELAI, while with MEL, there are some replicates which did not reach the success threshold.
However both MELAI and MEL produce similar results on the escape room.
More interestingly, on the three environments, the best and average initial task-performance (see second row of figure~\ref{fig:fitness}) of MELAI is above the one obtained by MEL. This shows that starting from a controller from the archive provides a better start for the learning algorithm.  

\begin{figure*}[t]
    \centering
    \includegraphics[width=0.32\textwidth]{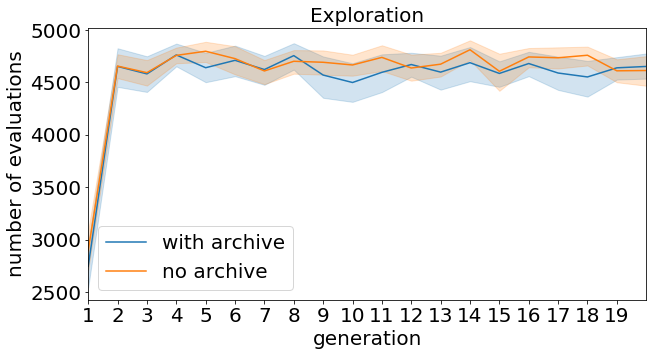}
    \includegraphics[width=0.32\textwidth]{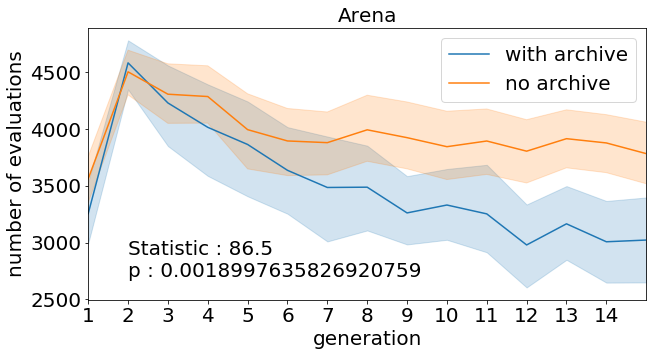}
    \includegraphics[width=0.32\textwidth]{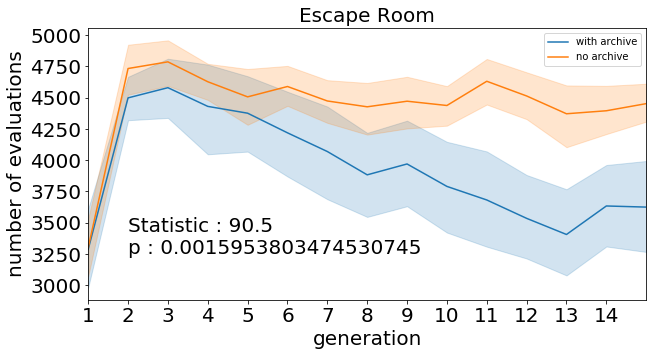}\\
    \caption{The number of evaluations per generation. The values are plotted for MEL and MELAI. These experiments have been conducted with a budget of 200 evaluations per body-plans and 20 generations for the exploration task and 15 for the photo-taxis task. The coloured areas correspond to the confidence interval around the mean. Difference between the distributions of the last generation is significant when a p-value and its critical value is indicated. The significant test is the Mann-Whitney U test.}
    \label{fig:ldne}
\end{figure*}

Another benefit of the controller archive is with respect to efficiency (figure \ref{fig:ldne}). On the photo-taxis task, the  total number of evaluations used per generation decreases over time for both algorithms (MEL and MELAI). The number of evaluations used by MELAI is fewer than MEL, and it aslo decreases faster (see first row of figure \ref{fig:ldne}). This dynamic does not appear in the exploration task because the learning algorithm does not have a target performance value for this task. So, when possible, transferring the controllers through the generation speeds up the learning. In other words, the learning process increases its efficiency over generations when the archive is used.

Moreover the difference between the initial and best performance values (learning delta) stays constant over the generations for both MELAI and MEL. As the archive allows the learning to start from better solution, MELAI can reach a better solution after learning and in a shorter time for the photo-taxis task.  Additional figures can be found in the supplementary material which plot the learning delta over the initial task-performance and the learning delta over the generations, and support the above interpretation.

\begin{figure*}[t]
    \centering
    \includegraphics[width=0.32\textwidth]{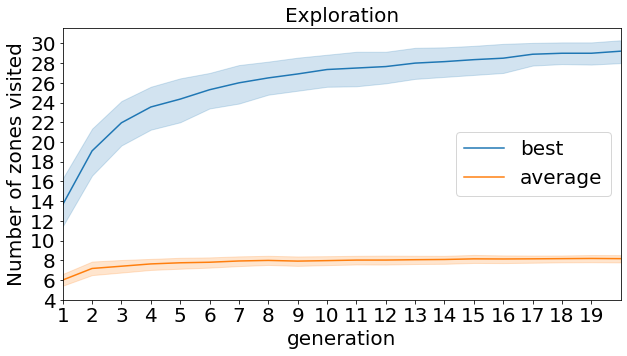}
    \includegraphics[width=0.32\textwidth]{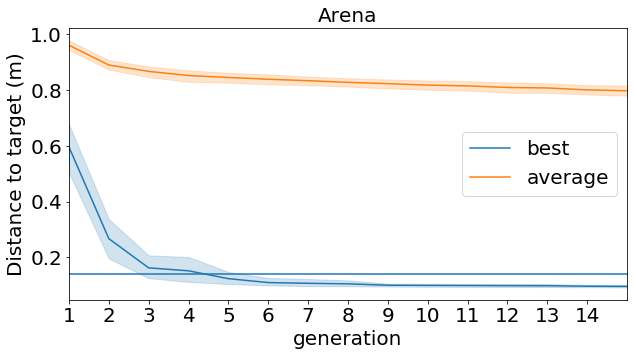} 
    \includegraphics[width=0.32\textwidth]{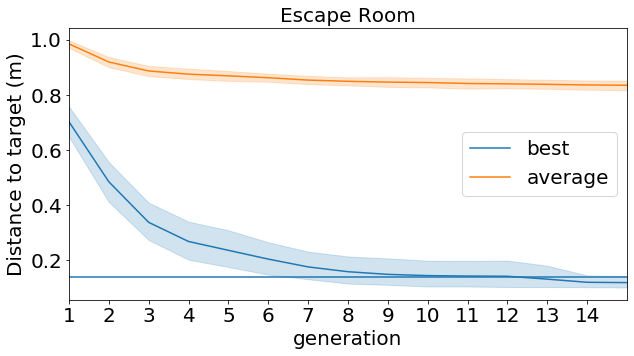} \\
    \includegraphics[width=0.32\textwidth]{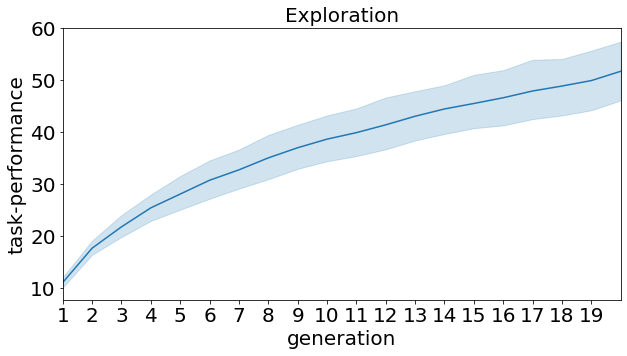}
    \includegraphics[width=0.32\textwidth]{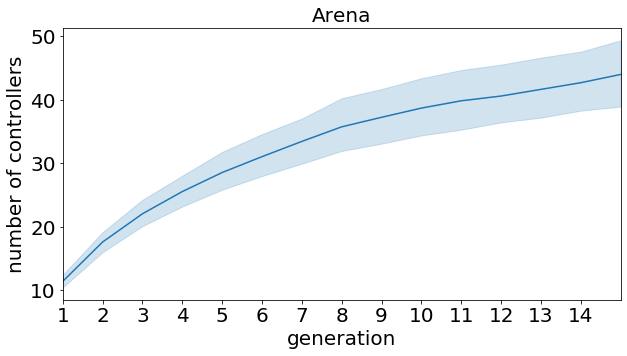}
    \includegraphics[width=0.32\textwidth]{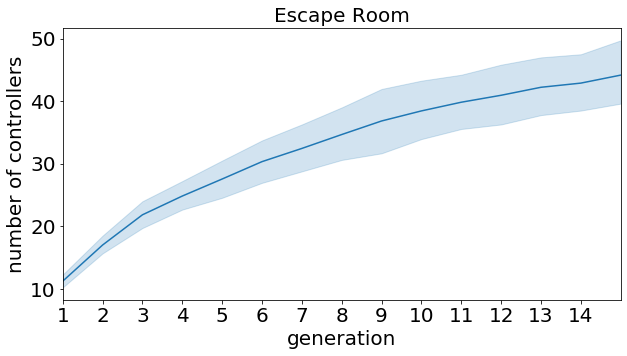}
    \caption{Plots of three metrics over the generations related to the controllers in the archive. First row: the average and best task-performance. Second row: the number of controllers in the archive. The coloured areas correspond to the confidence interval around the mean.}
    \label{fig:camet}
\end{figure*}

\subsection{Controller archive dynamics}

To analyse the dynamics associated with using the controller archive, two metrics related to the controllers stored in the archive are monitored: their average and best task-performance value (first row of figure~\ref{fig:camet}) and their number (second row of figure~\ref{fig:camet}). The number of controllers in the archive corresponds to the number of `types' of body-plan generated by the MEA (according to the 3-dimensional descriptor used). 

The average and best task-performance of the controllers stored in the archive follow the same dynamics as the best fitness of the population of body-plans (see figure~\ref{fig:fitness}). The \textit{average} task-performance value remains low as the controller archive retains controllers in cells that correspond to robots that have unsuitable body-plans for the tasks (e.g. no wheels).


The accumulation of controllers is shown in the second row of figure~\ref{fig:camet}. The controller archive accumulates controllers through the generations: some controllers are replaced over time by higher performing versions, while others may never be updated if the type of body-plan they belong to is not selected. The rate of increase however slows over time.

On the other hand the best task-performance in the archive corresponds to the best task-performance in the  population of body-plans (see figure~\ref{fig:fitness}).

\begin{figure*}
    \centering
    \includegraphics[width=0.7\linewidth]{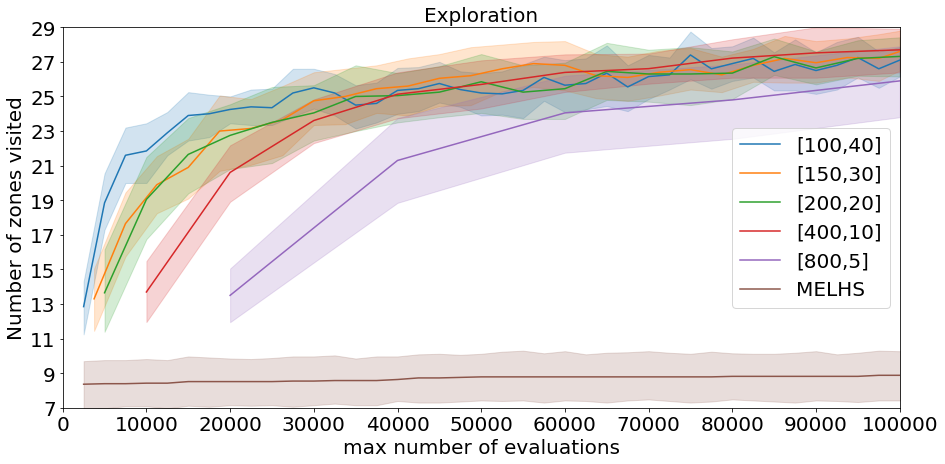} 
    \caption{Best fitness over the number of evaluations on the exploration task for six variants: [\textit{evaluations}, \textit{generations}] of [800,5],[400,10],[200,20], [150,30], and [100,40] and MELHS. The coloured areas correspond to the confidence interval around the mean.}
    \label{fig:evalvar}
\end{figure*}

\subsection{Influence of learning}

In addition to the controller archive, the combined evolution of morphology and learning of behaviours in MELAI leads to added complexity. To gain insight into the interaction between these two optimisation processes, MELAI was run with five different budgets ([800,5], [400,10], [200,20], [150,30], and [100,40]) which correspond respectively to [\textit{learning budget}, \textit{number of generation}]. All the variants have a population of 25 body-plans. Thus each variant tests a different total number of body-plans, for instance, the variant [400,10] tests 2500 body-plans. However, all the variants have the same total number of evaluations of 100000. These variants are compared with MELHS as baseline, conducted on the exploration task.

Figure~\ref{fig:evalvar} shows the best task-performances (number of zones visited) over the number of evaluations. As expected, the advantage of using learning is clear. The best individuals produced by MELHS are visiting on average between 8 and 9 zones while all the variants of MELAI reaches between 23 and 28 in average.  With MELHS, the quality of the solutions increases very slowly, starting from 8 and reaching barely 9 at the end.  However MELAI demonstrates an obvious learning curve for all variants tested. 

On the other hand, the difference between the different budget using MELAI is small. The variant [800,5] is sub-optimal comparing with the others. The variant [100,40] is the fastest to reach a satisfactory solution with an average 25 zones visited after around 15000 evaluations. Generally, reducing the learning budget speeds up the process. This is due to the generational aspect of MELAI: the smaller the learning budget, the faster the generations. Ultimately, apart from [800,5] all the variants converges to similar task-performances.

\begin{figure}
    \centering
    \includegraphics[width=0.8\linewidth]{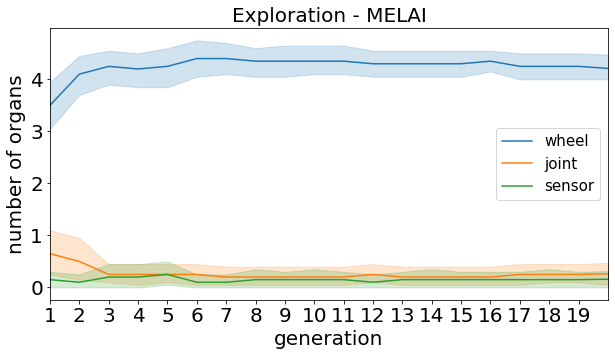}
    \includegraphics[width=0.8\linewidth]{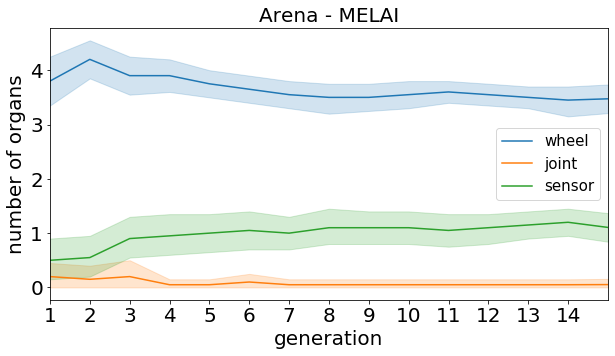}
    \includegraphics[width=0.8\linewidth]{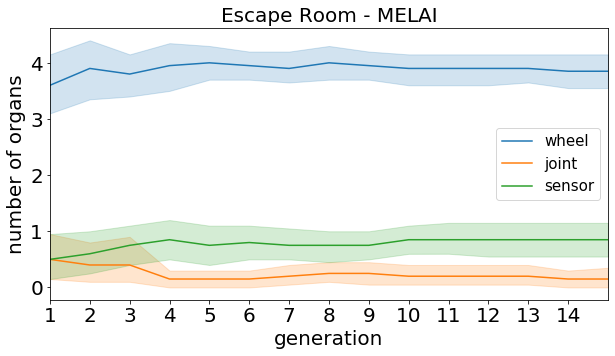}
    \caption{For each of the three environments, distribution of the number of wheels, joints and sensors over the best robots of each generation. The coloured areas correspond to the confidence interval and the solid curves to the central tendency.}
    \label{fig:nbr_organs}
\end{figure}

\subsection{Robot diversity}

Finally, figure~\ref{fig:nbr_organs} shows the distribution of number of wheels, joints and sensors of the robots with the highest task-performance over the generations. These show that the majority of successful robots have between three and five wheels for all the environments and both tasks. In contrast, the majority of successful robots have no joints. This is unsurprising as the three environments have a flat floor rendering joints unnecessary.  Interestingly, MELAI discovers that for photo-taxis at least one sensor is required, while for the exploration task the majority of the best solutions does not feature sensors. Indeed, to reach the target at three different locations in the photo-taxis, the robot  must have a sensor. On the contrary, blind robots can easily visits multiple zones.

\begin{figure}
    \centering
    \includegraphics[width=0.4\linewidth]{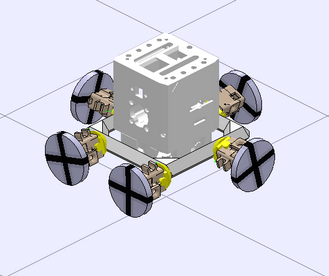}
    \includegraphics[width=0.4\linewidth]{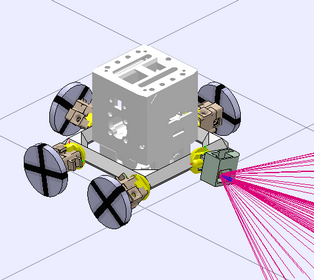}
    \caption{The most efficient design found by MELAI. On the left for the exploration task and on the right for the photo-taxis task.}
    \label{fig:robot_examples}
\end{figure}

This result is not surprising given that the main optimisation process in MELAI is an MEA. Evolution is most likely to proceed along the `easiest' path that enables it to maximise the fitness function. In this case, this corresponds to robots with only wheels. This type of robot is easier to control and therefore it is easier to learn a controller for them than robots with joints and sensors, even though the latter may be more efficient. Sensors emerge only if there are necessary like in the photo-taxis task.Therefore MELAI is able to produce different `types' of robots depending on the task. This is evidenced in additional plots provided in the Supplementary Material.
Pictures of the most efficient design found by MELAI are shown in figure~\ref{fig:robot_examples} as examples. Worth being noted the similarity and the minimalism of both body-plans. The similarity confirms the lack of diversity in the best robot generated. The minimalism shows that MELAI reduced the cost of production of such robots without an explicit objective.

\section{Discussion}\label{sec:disc}
Advancing previous work in the domain of body-brain evolution, we proposed a method to evolve robots in a rich morphological space that includes a variety of sensors and actuators and can realise skeletons with diverse forms and sizes. Hence, a considerably more diverse range of body-plans can be produced  in this space than in previous work that either uses modular systems \cite{miras2020evolving} or spaces in which the  components have common control mechanisms \cite{gupta2021embodied}. The richer space increases the likelihood that a controller produced via evolutionary operators will not match a new body-plan. 
Although using a generative (morphology-independent) encoding can address the inheritance issue, the time-complexity associated with these methods can be prohibitive when working with physical robots.


To address this we proposed the use of an external archive that stores a learned controller associated with a `type' of robot. As described in section \ref{sec:results}, we have demonstrated that the archive significantly improves the quality of the solution and the efficiency in evolving a body-plan capable of solving multiple tasks.  The role of  the archive and the various components of the framework used in leading to this result are discussed below.

Each cell of the controller archive stores the best controller learned during an individual lifetime. The archive thus represents a history of knowledge discovered in previous generations that can be passed to future generations. It therefore acts as a novel form of \textit{inheritance}. In the sense that it stores information learned during an individual lifetime, it shares characteristics with  Lamarckian artificial evolutionary systems \cite{jelisavcic2019lamarckian}. Note that the tuple defining a `type' is deliberately simple. However, it should be clear that many different body-plans can be mapped to each single cell, given that for any given combination of sensors and actuators, the robot-skeleton they are attached to can vary enormously in shape and size, and the configuration of components can also vary. Given this variation, it  might be expected that inheriting a controller of the correct `type' would not necessarily bring much benefit. However, it is clear from figure \ref{fig:fitness} that shows the best and initial task-performance, that starting from a controller from the archive brings a significant advantage. Interestingly, this suggests that there is some generalisation of controllers across a range of body-plans. The results shown in the second row of figure \ref{fig:fitness} show that inheriting from the archive bootstraps the learning process, and the size of this effect increases in magnitude as the generations progress. 

It would of course be possible to define each cell using a higher degree of granularity, although it is reasonable to  assume there is a balance to be struck in not making the archive too granular (which at the extreme would map every robot to an individual cell). Another way to approach this would be to store multiple controllers per cell, and either try them all, select one at random, or use a clustering or species system to select the most suitable one. Also, increasing the granularity of the archive requires defining an appropriate
 morphological descriptor. This is not an easy task. In one of our previous works \cite{buchanan2020bootstrapping}, different morphological descriptors were studied in the context of novelty search, designed to reflect different types of control (e.g. due to symmetry).
 The results shows that in fact the most simple descriptor that simply counted components was better in producing a greater diversity of body-plans.

Recall that the framework consists of two components: an MEA that learns body-plans and a learning algorithm based on an evolutionary strategy that learns controllers. The former is selected for its ability to explore a diverse space of plans and based on previous work \cite{buchanan2020bootstrapping}. The selection of NIP-ES as the learner is deliberate in that this algorithm demonstrates high \textit{exploration} capabilities. This is essential as the learner might have to start from scratch if no controller is available in the archive, or a selected controller might not be well adapted to a new body. In contrast previous work which has used learning as a mechanism to enhance a controller selected by evolution (e.g. \cite{miras2020evolving}) can afford to be much more exploitative. 

 
 Despite the diversity produced by MCME, the successful robots are all of the same `type', mostly wheels, a few sensors, and almost no joints. In fact MELAI, fall into the local optima of the robots for which it is easy to learn a controller. This issue is a common issue in evolutionary robotics and it is even accentuated by the joint optimisation of body-plan and controllers. In the work of Cheney et al. \cite{cheney2016difficulty}, this issue is explained by the fact that promising body-plans for which it takes longer to learn a controller are dropped by the EA. One of their latter works \cite{cheney2018scalable} proposed the morphological innovation protection mechanism. Each body-plan has an attribute corresponding to their age which increases at each generation. In addition of the task-performance, the EA selects the youngest body-plans and thus protect new body-plan which would needs more time to learn a controller. In MELAI, this solution could be implemented by attributing bigger budgets to the learning process for younger body-plans and decrease their budgets while they age. Another possible reason for the premature convergence in such local optimum is the generational aspect of the EA. At each generation the selection mechanism is applied to the whole population, thus, a high performing solution will often invade completely the population. The asynchronous parallel evolution (APE) proposed in the work of Gupta et al. \cite{gupta2021embodied} is a possible solution, indeed, they observe as a side effect, a diversity in the final high performing body-plans. APE performs tournament selection on small groups of four individuals asynchronously. So, a low performing robots could be preserved longer because it would not be always confronted to the high performing individuals.  
 
 Given the importance of the learning loop just discussed when jointly optimising body-plans and controllers, it is natural then to discuss how a computational budget should be balanced between the outer evolutionary loop and the inner learning loop. The results shown in figure \ref{fig:evalvar} shed some light on this by varying the budget assigned to the learning from 100 to 200  evaluations. The smaller learning budget delivers a faster bootstrap in both environments. It is also clear that using the archive results in lower variance, particularly noticeable when using the smallest learning budgets. However, the budget of 200 evaluations gives more consistent results over the environments and the two variants. Also, 200 evaluations is the necessary minimum budget to have NIP-ES to its full potential \cite{le2020sample} (see supplementary materials). So, the choice of budget is dependant on the learning algorithm used in MELAI. Also, it is worth remarking that the decision regarding how to split this budget is influenced by whether one is working in simulation or on physical robots: in simulation, generating a body-plan has negligible cost whereas in reality, producing a physical robot can take weeks \cite{liao2019data}. In contrast, evaluations are cheap in both environments hence this may influence the choice.
 
Finally, many design choices made in this work are aiming to match the present method with our physical system \cite{hale2020hardware}. In particular, NIPES and the controller archive reduce the number of evaluations needed to reach a satisfactory solution. Of course, the present framework cannot be applied directly on the real robotic platform. Indeed, on the photo-taxis task in the arena, MELAI needs about 15000 evaluations shared among 100 body-plans tested to reach a robots fulfilling the task. In future work, hybrid methods using both simulated and real robots will be investigated.

\section{Conclusion}\label{sec:concl}

This paper proposed a new framework MELAI for the joint optimisation of body-plans and controllers in a diverse and complex morphological space. The framework intertwines an evolutionary algorithm MCME for evolving body-plans and an evolution strategy NIP-ES for learning individual controllers. Its key novelty is in the use of an external archive for storing learned controllers for different `types' of robot. This acts as a means of transferring learned information between multiple generations, that is used to bootstrapping to learning mechanism. Hence it can be seen as a form  of non-genetic inheritance. It is shown to bring benefits with respect to efficiency, leading to increased rates and magnitude of learning over generations. Finally it provides new insights into the complex interactions between evolution and learning, adding to the growing amount of recent literature on this subject (e.g. \cite{gupta2021embodied,liao2019data,miras2020evolving}.

Looking ahead, the work provides a foundation for moving towards applying the framework to the evolution of robots in a hybrid system that mixes evolution in hardware and simulation: in such a space, increasing the efficiency of the evolutionary cycle is key for reasons that include time, cost of materials, and wear and tear on robotic parts.

\section*{Acknowledgment}
This work is funded by EPSRC ARE project, EP/R03561X, EP/R035733, EP/R035679, and the Vrije
610 Universiteit Amsterdam.

\section*{Appendix I: Algorithms Description}

\subsection{Matrix-based CPPN morpho-evolution}\label{sec:mcme}

\subsubsection{Body-plan decoding}

The body-plan decoding is a variation of the one proposed in the work of Buchanan \textit{et al}.\cite{buchanan2020bootstrapping}. The CPPN genome in this paper has four inputs and five outputs. Three inputs represent the x, y and z coordinates of a cell in a 3D matrix to queried and a fourth input represents the distance from the cell of the matrix to the centre of the matrix. Each of the outputs defines the presence or absence of the skeleton and/or component of each type.

The genome decoding takes place in four steps:

\begin{enumerate}
    \item All the cells in the 3D matrix are queried to generate the skeleton of the robot.
    \item A repair mechanism makes changes to the skeleton to meet the printing restrictions. Some of the restrictions include: make sure there is only one piece of skeleton and the skeleton is connected to the base of the head organ.
    \item All the cells on the surface of the skeleton are queried to generate the organs. For this, the four outputs of the organs are taken. The output with the highest value defines the organ to be place on the cell.
    \item A second repair mechanism removes colliding organs. 
\end{enumerate}

\begin{figure}[h]
\begin{center}\includegraphics[width=0.8\linewidth]{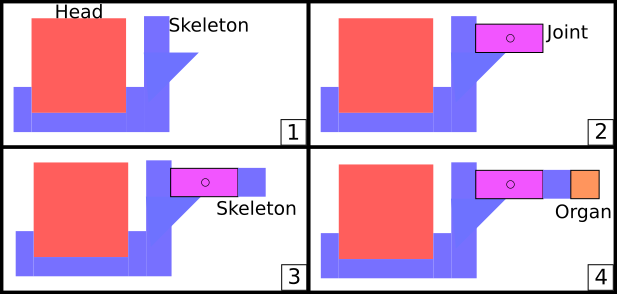} 
\end{center}
\caption{Generation of multi-segmented robots. (1) The main skeleton is generated first. (2) A joint is place on the surface of one of the voxels. (3) a cuboid skeleton with 4 cm side is generated at the other end of the joint. (4) The CPPN is queried to generate components at each side of the cuboid.} 
\label{fig:bodyplan}
\end{figure}

The decoding used in this paper has the additional feature of generating multi-segmented robots and it works as follows. The position of each skeleton voxel is queried in CPPN (Figure~\ref{fig:bodyplan}.1). If the component generated is a joint (Figure~\ref{fig:bodyplan}.2) then a cuboid skeleton is generated at the other end of the joint (Figure~\ref{fig:bodyplan}.3), The position of each face of cuboid is queried to the same CPPN and components are generated (Figure~\ref{fig:bodyplan}.4). The work of Hale \textit{et al}.\cite{hale2020hardware} described how the physical multi-segmented robot are assembled in the robot fabricator.

The algorithm evolving the CPPN is the neuro-evolution of augmenting topology (NEAT) \cite{stanley2002evolving} which is a generational EA using a generative encoding to evolve both the topology and the weights of the network. In this work, we use the implementation of NEAT from the MultiNEAT library\footnote{\url{http://www.multineat.com/}}.

\subsubsection{Manufacturability restrictions}

Each component in the body-plan has to meet the same manufacturability criteria introduced in the work of Buchanan \textit{et al}.\cite{buchanan2020bootstrapping}. If an component fails any of the manufacturability tests then the component is removed from the final body-plan phenotype.

The physical head organ has 8 electrical connections for components, therefore only up to 8 active components can be connected to head skeleton at any time. The joints offer the option to electrically daisy chain one more active component. In total, a body-plan can have up to 16 active components.  

The size of the skeleton connected to the head component can be as big as 23cm x 23 cm x 23 cm. 
\begin{figure}[h]
\begin{center}
\includegraphics[width=\linewidth]{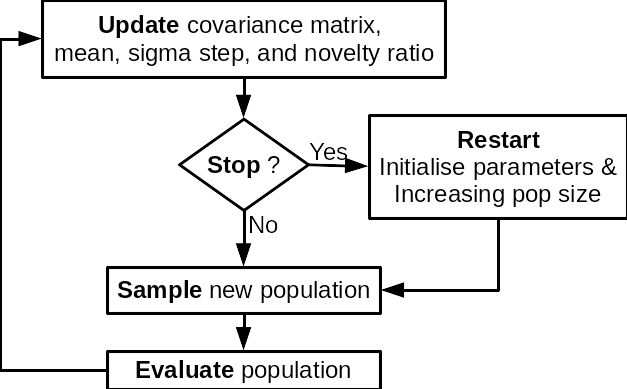} 
\end{center}
\caption{Diagram describing NIP-ES algorithm} 
\label{fig:nipes}
\end{figure}

\subsection{NIP-ES}\label{sec:nipes}

The novelty-driven increasing population evolutionary strategies (NIP-ES) is a learning algorithm introduced in one of our previous works \cite{le2020sample}. Primarily, this algorithm was designed to find a solution using as few evaluations as possible.  NIP-ES is a custom version of the increasing population co-variance matrix adaptation evolutionary strategies (IPOP-CMA-ES) proposed by Auger and Hansen \cite{auger2005restart}.  NIP-ES is an algorithm of the CMA-ES family \cite{hansen2006cma,hansen2016cma} in which a multivariate normal distribution (MVND) is used to sample a set of solutions to be evaluated. This set of solutions is equivalent to the population of an EA. 

An iteration of a CMA-ES consists in three steps (see figure~\ref{fig:nipes}): 

\begin{itemize}
    \item[(1)] \textit{update} the co-variance matrix and mean of the MVND; 
    \item[(2)] \textit{sample} a new population;
    \item[(3)] \textit{evaluate} the population.
\end{itemize}

The magnitude of change of the co-variance matrix is controlled by a parameter called sigma step. This parameter is similar to the learning rate in reinforcement learning. From an initial value ($\sigma_0$), the sigma step decreases at each iteration and so, the adaptation of co-variance matrix slows down. Among, the different updates function of the co-variance matrix existing in the CMA-ES family, naturally, NIP-ES uses the one of IPOP-CMA-ES implemented in libcmaes\footnote{\url{https://github.com/CMA-ES/libcmaes}} and used in the benchmark of CMA-ES by Hansen \cite{hansen2009benchmarking}.

In IPOP-CMA-ES can restart under certain conditions. After a restart, the MVND's parameters along with the sigma step are reinitialised and the population size is increased by a factor two. As shown in figure~\ref{fig:nipes}, NIP-ES features the same restart mechanism, but differs in the stopping conditions. Two conditions can trigger the restart of NIP-ES:

\begin{itemize}
    \item \textit{Best task-performance stagnation} : Over a window of 20 iterations if the standard deviation of the best task-performance values are below a threshold ($\tau_1$);
    \item \textit{Low behavioural diversity} : If the standard deviation of the populations behavioural descriptors are below a threshold ($\tau_2$). In this paper, the behavioural descriptor of an individual is its final position in the arena.
\end{itemize}

These two conditions aim at detecting when the algorithm get stuck in a local optimum.

Finally, NIP-ES' fitness function is a weighted sum of two objectives: the task-performance value and the behavioural novelty score (see equation~\ref{eq:objnsr}). The novelty score measures how much the behaviour of an individual is new in comparison with the other individual in the population and past individuals stored in an archive \cite{lehman2011abandoning}. The novelty score is computed by averaging the distances between the individual and its 15 nearest neighbours in the population and the archive. The archive of past individuals is updated at each iteration by adding randomly a part of the population and individuals with a novelty score above a threshold. 

\begin{equation}
F = \eta*S + (1-\eta)*r  
\label{eq:objnsr}
\end{equation}

The objectives are weighted with a novelty ratio ($\eta$), the novelty score ($S$) is multiplied by the novelty ratio ($\eta$) and the task-performance value ($r$) by the opposite novelty ratio (see equation~\ref{eq:objnsr}). The novelty ratio starts at one and then decreases by a fix decrements ($\eta_d$). When the algorithm restart the novelty ratio is reinitialized at one. NIP-ES starts with a pure exploratory behaviour to slowly transitions to a exploitative behaviours. 

In the context of this work, NIP-ES has three stopping conditions:
\begin{itemize}
    \item if an individual achieves a task-performance value above a success threshold ($\tau_S$);
    \item if the maximum budget of evaluations is reached. The budget can be exceeded when the size of the last population is greater than the number of evaluations remaining. 
    \item if after a trial period of 50 iterations is passed with reaching the minimal task-performance value. 
\end{itemize}

This last condition was introduced in order for MELAI to detect when a body-plan does not have the minimum capability required to solve the task. In a navigation task, it simply  detects whether the robot has moved after a certain simulation duration.

NIP-ES is constituted of cycles which first explores the space to locate new solutions, then exploits the most promising ones. And after each restart, the exploration power of the algorithm increases by doubling the population's size. By starting with a small population and only increasing it if necessary, NIP-ES tends to use the minimum necessary number of evaluations \cite{le2020sample}.  With the hyper-parameters used in this paper, these cycles are approximately 20 iterations. Therefore with a starting population of 10, the minimum budget needed in order for NIP-ES to operate at its full potential is 200. This estimation results from the fact that the first stopping criterion can be triggered every 20 iterations and the second criterion is the least probable to be triggered. The hyper-parameters and their values for each experiments of this article are listed in table~\ref{tab:hpnipes}.

\begin{table}
\caption{hyper-parameters of NIP-ES}
\centering
\begin{tabular}{|l|l|}
\hline
Initial sigma step ($\sigma_0$) & 1 \\ \hline
Initial population size & 10 \\ \hline
Initial novelty ratio ($\eta_0$) & 1 \\ \hline
Novelty ratio decrements ($\eta_d$) & 0.05\\ \hline
Sparseness number of nearest neighbors & 15 \\ \hline
Novelty threshold to add to archive & 0.9 \\ \hline
Probability to add to archive & 0.4 \\ \hline
Simulation time & 60 seconds \\ \hline
Best task-performance stagnation threshold ($\tau_1$) & 0.05 \\ \hline
Low behavioural diversity threshold ($\tau_2$) & 0.1 \\ \hline
Trial period (number of iteration) & 50 \\ \hline
Success threshold ($\tau_S$) & 0.95 \\ \hline
\end{tabular}
\label{tab:hpnipes}
\end{table}

\newpage

\section*{Appendix II: Complementary Plots and Figures}

\subsection{Elman Network}

\begin{figure}[h]
\begin{center}
\includegraphics[width=\linewidth]{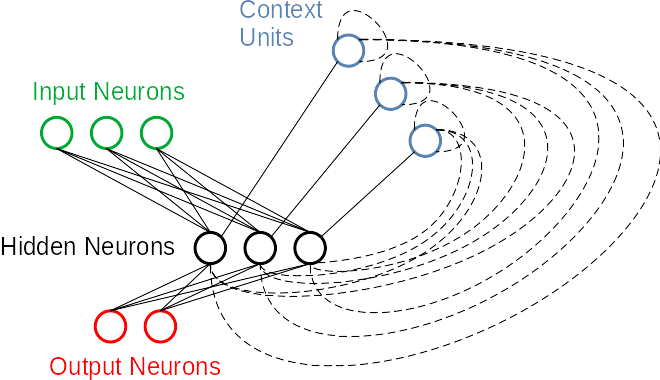} 
\end{center}
\caption{Diagram describing the Elman network structure. Solid lines correspond to forward connections and dashed lines to backward connections.} 
\label{fig:elmannet}
\end{figure}
\FloatBarrier

\subsection{Learning Delta over the initial task-performance}

\begin{figure}[h]
\begin{center}
\includegraphics[width=\linewidth]{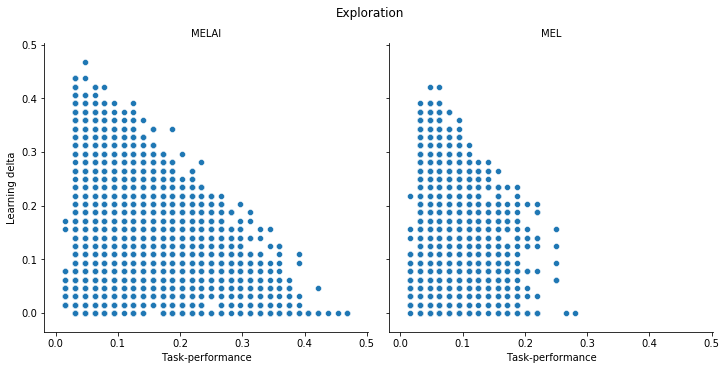} \\
\includegraphics[width=\linewidth]{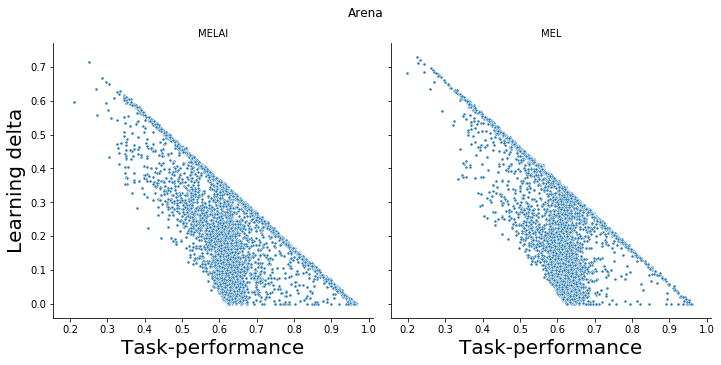} \\
\includegraphics[width=\linewidth]{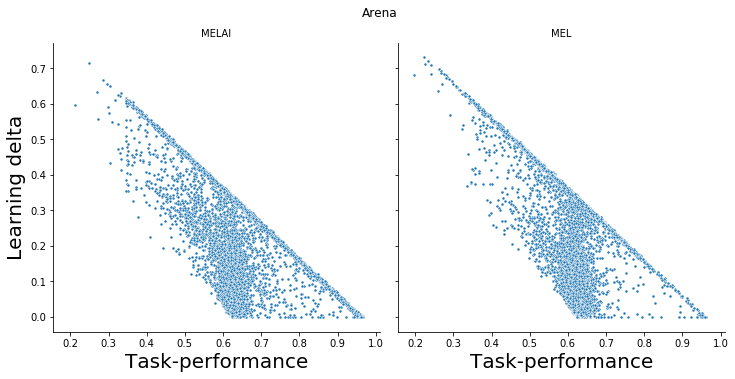} 
\vspace{-10mm}
\end{center}
\caption{Scatter plots of the learning delta over initial task-performance. All the values are given with the normalised task-performance} 
\label{fig:ifxld}
\end{figure}

The \textit{learning delta} is defined as the difference between the the initial and best task-performance. Figure~\ref{fig:ifxld} shows that the learning delta decreases linearly when the initial task-performance increases. This implies that starting from a better initial solution simply makes it easier to reach a better solution.
These plots suggest that there is no clear benefit gained from the archive. Only, MELAI reaches higher initial task-performance on the exploration task and on the photo-taxis task, the distribution for MEL is more packed around the mean (0.6 of initial task-performance). 

\begin{figure}[h]
\begin{center}
\includegraphics[width=\linewidth]{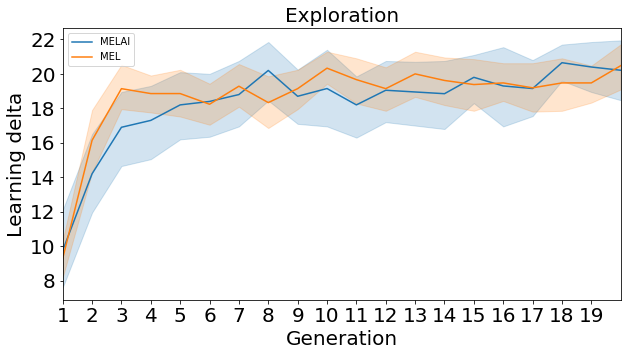}\\
\includegraphics[width=\linewidth]{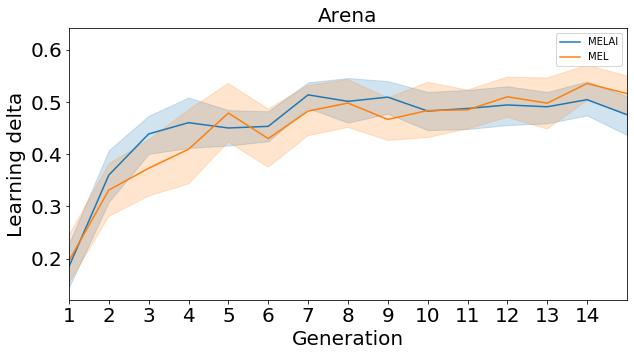}\\
\includegraphics[width=\linewidth]{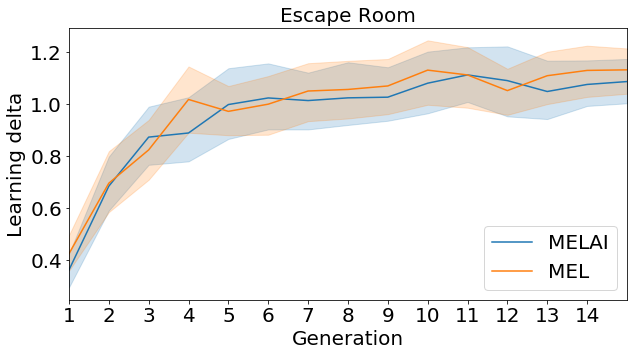} 
\vspace{-5mm}
\end{center}
\caption{Plots of the learning delta over the generations.} 
\label{fig:ifxld}
\end{figure}
\FloatBarrier
\newpage
\subsection{Type of robots produced by MELAI}

\begin{figure}[h]
\begin{center}
\includegraphics[width=0.95\linewidth]{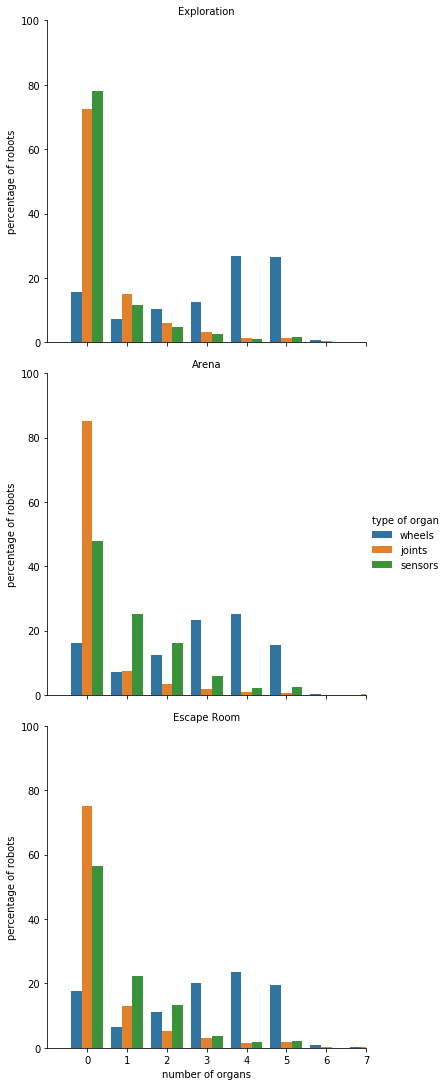} 
\vspace{-10mm}
\end{center}
\caption{Type of components distributions over all the robots produced by MELAI} 
\label{fig:elmannet}
\end{figure}

 \newpage
 
\bibliographystyle{IEEEtran}
\bibliography{bib}

\ifCLASSOPTIONcaptionsoff
  \newpage
\fi



%

\end{document}